\documentclass[letterpaper, 10 pt, conference]{ieeeconf}  

\IEEEoverridecommandlockouts                              
\usepackage{graphics} 
\usepackage{amsmath} 

\usepackage{mathtools,amsthm}
\usepackage{amssymb}  
\usepackage{graphicx}
\usepackage[table,dvipsnames]{xcolor}
\usepackage{placeins}
\usepackage[font={small}]{caption}
\usepackage{subcaption}
\usepackage{verbatim}
\usepackage{hyperref}
\usepackage{epstopdf}
\usepackage{afterpage}
\usepackage{siunitx}

\usepackage{multirow}
\usepackage{hhline}

\DeclareMathOperator*{\argmin}{arg\ min.}

\newcommand{\mytheorem}[2]{%
\newtheorem{t#2}{#1}%
\newenvironment{#2}{\begin{t#2}}{\end{t#2}}}

\newcommand{\myremark}[2]{%
\newtheorem{t#2}{#1}[section]%
\newenvironment{#2}{\begin{t#2}}{\end{t#2}}}

\theoremstyle{plain}

\mytheorem{Theorem}{theorem}
\mytheorem{Lemma}{lemma}
\mytheorem{Proposition}{proposition}
\mytheorem{Corollary}{corollary}
\mytheorem{Assumption}{assumption}
\mytheorem{Definition}{definition}
\mytheorem{Property}{property}
\myremark{Remark}{ownremark}

\newcommand{\rmf}{{\rm f}}      

\author{Francisco Javier Andrade Chavez$^{1,2}$, Gabriele Nava${^2}$, Silvio Traversaro$^{2}$, Francesco Nori$^{2}$ and Daniele Pucci$^{2}$
\thanks{
This project has received funding from the European Union’s Horizon
2020 research and innovation programme under grant agreement No. 731540
(An.Dy).} 
\thanks{$^{1}$ Francisco Javier Andrade Chavez with CVU 468287 receives support from the National Council of Science and Technology (CONACYT) in Mexico}
\thanks{$^{2}$ All authors belong to the Istituto Italiano di Tecnologia, iCub Facility department, Genova, Italy.
       Emails: {\tt\footnotesize ing.andrade.francisco@gmail.com}, {\tt\footnotesize gabriele.nava@iit.it},  {\tt\footnotesize silvio.traversaro@iit.it},  {\tt\footnotesize francesco.nori@iit.it},  {\tt\footnotesize daniele.pucci@iit.it}}
\thanks{Digital Object Identifier (DOI): see top of this page.}
}

\title{Model Based In Situ Calibration with Temperature compensation of 6 axis Force Torque Sensors}

\begin{document}

\maketitle

\begin{abstract}
It is  well  known  that  sensors  using  strain  gauges  have a  potential  dependency  on  temperature. This creates temperature drift in the measurements of six axis force torque sensors (F/T).  The temperature drift can  be  considerable  if  an  experiment is  long  or  the  environmental  conditions  are  different  from when  the  calibration  of  the  sensor  was  performed. Other \textit{in situ} methods disregard the effect of temperature on the sensor measurements. Experiments performed using the humanoid robot platform  iCub show that the effect of temperature is relevant. The model based \textit{in situ} calibration of six axis force torque sensors method is extended to perform temperature compensation.
\end{abstract}


\section{INTRODUCTION}
Six  axis  force  torque  (F/T)  sensors have  been  used for years in robotic systems~[1]. They have not being able to be fully exploited in floating base robots due to unreliability of the sensors. This unreliability arises from the scenarios in which floating base robots are expected to be used. A clear example of this can be taken from the DARPA Robotics Challenge \cite{DRC-what-happened}.

 In standard operating conditions, a decrease in the effectiveness of the calibration may occur in months. Leading companies for F/T sensors \cite{atimanual,kms40manual} recommend to calibrate the sensors at least once a year. The calibration done by the manufacturer usually implies that the sensor must be unmounted, sent back to them and then mounted again.
 
 F/T sensors are prone to change performance once mounted in a mechanical structure such as a robot \cite{InSituAcc,insituFTcalibration}. Different methods have been developed to re-calibrate the sensors once mounted.These \textit{in situ} methods allow to perform the calibration in the sensor's final destination, avoiding the decrease in performance that arise from mounting and removing the sensors from its working structure. The relevance of calibrating \textit{in situ} has become evident, making \textit{in situ} calibration part of the service provided by F/T sensor companies \cite{kistlerCalibrationBrochure}.
 
 Is well known that that sensors using strain gauges have a potential dependency on temperature \cite{helmick2006comparison}, which creates temperature drift. It can be considerable if an experiment is long or the environmental conditions are different from when the calibration was performed \cite{sun2016temperature}. A common strategy in F/T sensors to reduce the effect of drift is to remove the bias just before a change in the load is expected. In floating base robots, this is not practical since most of the time the sensors themselves are used to detect the contact so the time of collision is not known \textit{a priori}. Besides, the main function of the sensors is to measure the actual force applied or received by the robot. In a scenario in which the robot is already in contact with a surface, removing the bias will make the value of the measured F/T incorrect.  For this reasons, being able to minimize the effect of drift in the F/T sensors can improve the reliability of the sensor in floating base robots.
 
  To the best of our knowledge, the first F/T sensor \textit{in situ} calibration method exploited the topology of a specific kind of manipulators equipped with joint torque sensors. These were then leveraged during the estimation \cite{shimanoroth}. The temperature was not considered.
Another \textit{in situ} calibration method for F/T sensors can be found in \cite{roozbahani2013novel}. But, the use of supplementary already-calibrated force-torque/pressure sensors, impairs this method since those sensors are prone to be affected by the mounting procedure, propagating the error from sensor to sensor. Another disadvantage is depending on the availability of another sensor. Temperature is carefully regulated during calibration, but changes in temperature on the working conditions are not accounted for. 
Some methods rely on adding other external sensors such as accelerometers to obtain a ground truth \cite{InSituAcc}. This translates the source of error to the accuracy of the accelerometers and measurement of the transformation matrix between the sensor frames. Other methods exploit the encoders and the model of the robot to provide the reference forces and torques \cite{insituFTcalibration}.  Both of these methods disregard the effect of drift in the sensor by assuming experiments are short enough. Experiments performed using the humanoid robot platform iCub show that the effect of temperature is relevant.

The aim of this paper is to extend the model based \textit{in situ} calibration of six axis force torque sensors method \cite{insituFTcalibration} to account for the temperature. 

The paper structure is as follows: the notation can be found in Section \ref{sec:background} as well as a description of the current model based \textit{in situ} calibration of six axis force torque sensors method and the effect of temperature on the F/T measurements. The problem  statement and the contribution of this paper are described in Section \ref{sec:method}. Section \ref{sec:experiments} details the experiments. Results are shown in Section \ref{sec:results} and the conclusions can be found in Section \ref{sec:conclusions} .

\section{BACKGROUND}
\label{sec:background}
\subsection{Notation}
The following notation is used throughout the paper.
\begin{itemize}
 \item The Euclidean norm of either a vector or a matrix of real numbers is denoted by $\left\| \cdot \right\|$.
 \item $n$ is the number of data points in a data set.
\item $ \rmf = \begin{bsmallmatrix}  f \\  \tau \end{bsmallmatrix}$ are a 6D force $\rmf$.
\item $\overline{r} \in \mathbb{R}^{n,1}$ is a column vector 
\item $R= [\overline{r}_{1}, \overline{r}_{2},...., \overline{r}_{j} ]$, is a matrix $R  \in \mathbb{R}^{n,j}$ 
\item $I_n \in \mathbb{R}^{n \times n}$ denotes the identity matrix of dimension~$n$
\item $0_n \in \mathbb{R}^n\times1$ denotes the zero column vector of dimension~$n$
\item $0_{n \times m} \in \mathbb{R}^{n \times m}$ denotes the zero matrix of dimension~$n \times m$.
\item Given $A \in \mathbb{R}^{n \times m}$ and $B \in \mathbb{R}^{p \times q}$, we denote with $\otimes$ the Kronecker product $A \otimes B \in \mathbb{R}^{np \times mq}$.
\item Given $X \in \mathbb{R}^{m \times n}$, $\text{vec}(X) \in \mathbb{R}^{nm}$ denotes the column vector obtained by stacking the columns of the matrix~$X$. In view of the definition of $\text{vec}(\cdot)$, it follows that \begin{equation}\label{eq:kroneckerVec} \text{vec}(AXB) = \left( B^{\top} \otimes A \right) \text{vec}(X).\end{equation}
\end{itemize}

\subsection{Previous sensor model}
The previous model considered the sensor as linear with the following form:
\begin{equation}
\rmf =Cr+o \label{eq:lin}
\end{equation}
where $\rmf \in \mathbb{R}^6$  are the 6D forces, $C \in \mathbb{R}^{6 \times 6}$ is the calibration matrix, $r \in \mathbb{R}^6$ are the raw measurements and $o \in \mathbb{R}^6$ is the offset. The calibration matrix $C$ and the offset $o$ are the variables to be estimated.
The offset was estimated separately using 2 different strategies. The \textit{in situ} offset estimation\cite{InSituAcc} and the centralized offset removal. Abusing notation the inputs of the problem are formulated as:
\begin{equation}
\hat{r}_i= \begin{cases} 
      r_i-o_r & \text{\textit{in situ} offset estimation} \\
      r_i-\mu_r & \text{centralized offset removal}
   \end{cases}
\end{equation}
\begin{equation}
\hat{\rmf}_i = \begin{cases} 
      \rmf_i & \text{in situ offset estimation} \\
      \rmf_i-\mu_\rmf  & \text{centralized offset removal}
   \end{cases}
\end{equation}
Where $r_i$ is a raw measurement coming from the sensor, $\rmf_i$ a the 6D force estimated using the model, $\mu_r$ and $\mu_f $ are the mean values of the raw measurements and the estimated 6D forces respectively, $\hat{r}_i$ and $\hat{\rmf}_i$ are the data used to solve the model based \textit{in situ} calibration problem in the following form:
\begin{equation}
\label{eq:modelInSitu}
 C^{*} =   \argmin_{C \in \mathbb{R}^{6\times6}}\ ~ \frac{1}{n}\sum_{i = 1}^n \left\|\hat{\rmf} _{i} - C\hat{r_i} \right\|^2 \\+\lambda\left\|C-C_w\right\|^2,
\end{equation}
where $\lambda$ is used to decide how much to penalize the regularization term, $C_w \in \mathbb{R}^{6 \times 6}$ is the calibration matrix provided by the manufacturer by calibrating the sensor on a \emph{workbench}. As such we will refer to it as \emph{Workbench} matrix. The regularization is added in order to try to keep the calibration matrix close to the calibration obtained using the workbench with an improved performance once the sensor is mounted on the mechanical structure. 
\begin{figure}[hbt!]
    \centering
      \includegraphics[width=\linewidth]{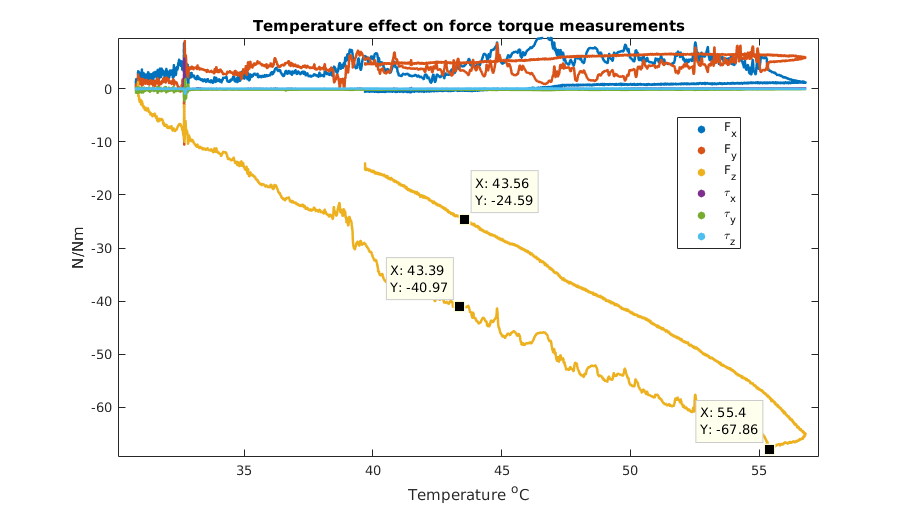}
        \caption{Temperature effect on the F/T measurements}
         \label{fig:temperatureEffect}
\end{figure}
\begin{figure}[hbt!]
    \centering
      \includegraphics[width=\linewidth]{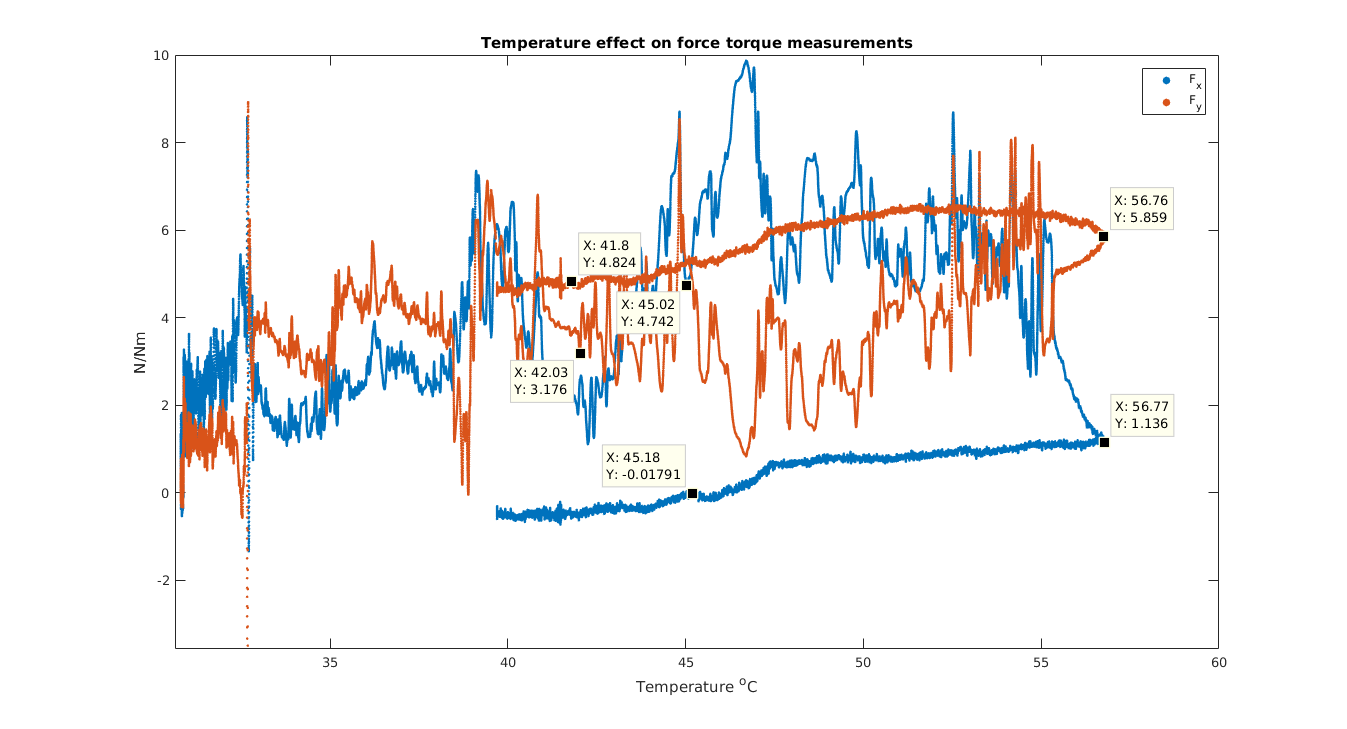}
        \caption{Temperature effect on the F/T measurements}
         \label{fig:temperatureOnForces}
\end{figure}
\begin{figure}[hbt!]
    \centering
      \includegraphics[width=\linewidth]{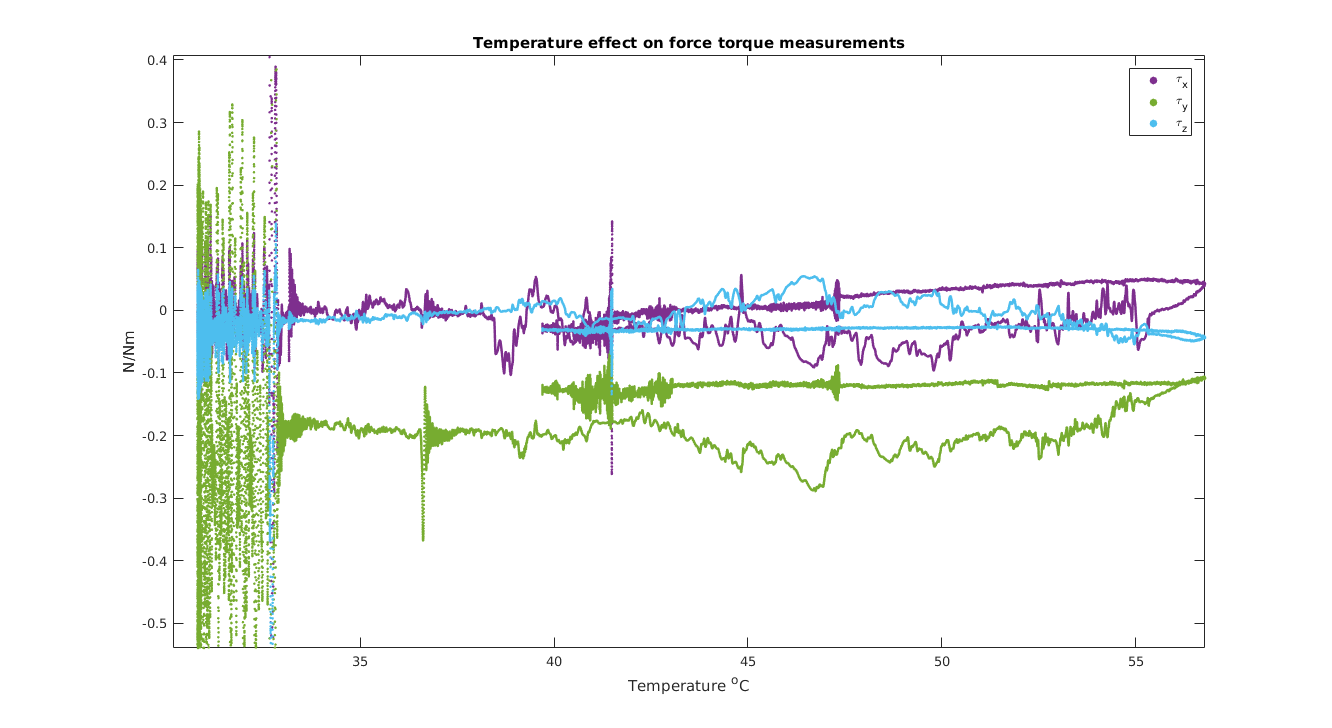}
        \caption{Temperature effect on the F/T measurements}
         \label{fig:temperatureOnTorques}
\end{figure}
\subsection{The temperature effect} \label{sec:temperatureEffect}
 Recently the custom made 6 axis F/T sensor (FTsense)\cite{IITsensors} have been modified to include a temperature sensor. This allows to study the effect of temperature in the measurements. A common solution to compensate temperature effects when using strain gauges is to use the Wheatstone bridge circuit to compensate for temperature \cite{hoffmann1974applying}. But this method is effective to compensate for temperature only if all strain gauges are  subjected to the same temperature change. Given the dimensions and arrangement of the strain gauges inside the FTsense, applying this method to compensate for temperature was not feasible when the sensor was designed. 
As a result, temperature has a considerable effect on the F/T measurements as shown in Fig. \ref{fig:temperatureEffect}.  In this experiment, a heat gun was used to heat a F/T sensor while measuring the load of a 33kg robot. The temperature effect is most visible on the z-axis, which is the one receiving most of the load. It seems this effect is close to a linear behavior. The temperature does affect the other axis as shown in Fig. \ref{fig:temperatureOnForces} and Fig. \ref{fig:temperatureOnTorques}. In these figures the bias is removed to better showcase the effect of temperature in the forces and torques. The temperature seems to be more effect in the forces than in the torques. The observed vibration while heating up was induced by the air coming from the heat gun. The effect of hysteresis can also be appreciated in the figures.

\section{METHODOLOGY}
\label{sec:method}
\subsection{Problem Statement}
The temperature creates a drift that seems to have a linear behavior. While using the robot the main heat source are the motors. As such, it is a safe assumption that the temperature drift, while using the robot, will be due to the temperature rising and not decreasing. With this in mind, hysteresis in the temperature drift can be ignored.
We assume that temperature is the main cause of drift so the offset can be consider constant.
Taking all this into consideration, the model is extended to account for temperature as follows:
\begin{equation}
\rmf =Cr+o+C_{t}t \label{eq:lin}
\end{equation}
where $C_{t}$ are the temperature calibration coefficients and $t$ is the temperature value.
In this case the problem is not only to estimate the calibration matrix $C$ and the offset $o$, but also $C_{t}$ which will account for the temperature changes in the sensor. Similar to \cite{insituFTcalibration}, we estimate the offset separately and include some regularization parameters to penalize the difference with respect to $C_w$. Currently the temperature when the sensor was calibrated is not provided. The final form of the problem is then:
\begin{equation}
\label{eq:modelInSituWthTemp}
 C^{*},C_t^{*} = \argmin_{C \in \mathbb{R}^{6\times6}}\ ~ \frac{1}{N}\sum_{i = 1}^N \left\|\hat{\rmf} _{i} - (C\hat{r_i}+C_{t}t)\right\|^2 \\+\lambda\left\|C-C_w\right\|^2.
\end{equation}
\subsection{Adding the temperature as a linear variable}
\label{sec:addLinearVariable}
Even if the 6 axis can be considered independent problems and solved individually, we solve them all together for convenience purposes. This is performed doing the following steps:
\begin{itemize}
\item {Consider the Matrix form of the least squares
\begin{equation}
     \left\|F^{\top} - CR^{\top}\right\|^2 +\lambda\left\|C-C_w\right\|^2 \label{eq:matrixForm},
\end{equation}
where $F^{\top} \in \mathbb{R}^{6,n}$ is the matrix with the reference 6D forces where each columns is $\hat{\rmf}_i$, $R^{\top} \in \mathbb{R}^{6,n}$  where each column is $\hat{r}_i$.}
\item {If we consider that $CR^{\top}=I_6CR^{\top}$ then, using the Kronecker property mentioned in eq. \ref{eq:kroneckerVec}, we can put eq. \ref{eq:matrixForm} in the column vectorized form:
\begin{equation}
    \left\| vec(F^{\top}) - (R \otimes I_6) vec(C) \right\|^2 +\lambda\left\|vec(C)-vec(C_w)\right\|^2.
\end{equation}
}
\item It is straight forward to show that the solution to the vectorized form of this problem is given by
\begin{equation}
    C^*=(K_R^{\top}K_R + \lambda I_{6*6})^{-1} (K_R^{\top} vec(F^{\top}) + \lambda vec(C_w)),
\end{equation}
where $K_R=  (R \otimes I_6)$. It is important to notice that the size of $I$ multiplying lambda should match the length of $vec(C_w)$ which is  $a*\rho$, where $a$ is the number of axis (6) and $\rho$ is the number of raw signals (6).
\end{itemize}
Given that $C\hat{r}_i + C_t t= \begin{bsmallmatrix}C , C_t\end{bsmallmatrix} \begin{bsmallmatrix}
\hat{r}_i \\ t
\end{bsmallmatrix} $ adding temperature can be considered adding an extra raw signal to the previous mentioned solution. It comes down to:
\begin{itemize}
\item Augment the raw measurements matrix $R$ with the temperature value $R_a= [R, \overline{t}], \overline{t} \in \mathbb{R}^{n,1}$
 , in $R$ each column has all the raw measurements of a given raw signal.
    \item Augment the workbench calibration matrix by including the coefficients regarding temperature $C_{w_a}=[C_w, C_{t_w}]$, where $C_{t_w}$ refers to the temperature at the time of calibration which is currently not available, so is set to $0_6$.
    \item Since $C_{w_a} \in \mathbb{R}^{6,6+1}$ this should be reflected in $L=\lambda*I_{6*(6+1)}$, since the workbench temperature coefficients $C_{t_w}$ are not provided, it is convenient to set the last $a$ values in the diagonal($L)$ to 0. This reflects the fact that we do not want to influence the coefficients of temperature with any previous information.
    \item The final form of the solution is
    \begin{equation}
         \begin{bsmallmatrix}C , C_t\end{bsmallmatrix}^*=(K_{R_a}^{\top}K_{R_a} + L)^{-1} (K_{R_a}^{\top} vec(F^{\top}) + L vec(C_{w_a}))
    \end{equation}
\end{itemize}
This allows to easily expand the solution to $m$ number of extra linear variables, in the case of temperature is just 1.
\subsection{Estimation Types}
\label{sec:estimationTypes}
Each strategy of offset estimation will be considered an estimation type. Including temperature or not in the estimation will be also considered different estimation types. Resulting in the following 4 estimation types:
\begin{itemize}
    \item Sphere with no temperature (\textbf{SnT}): Refers to the fact that the \textit{in situ} offset removal is obtained by expecting a sphere in the force space when generating circular motions. No temperature considered.
    \item Centralized with no temperature (\textbf{CnT}): Refers to the centralized offset removal method without considering temperature.
    \item Sphere with temperature (\textbf{SwT}): Refers to including temperature into the sphere type.
    \item Centralized with temperature (\textbf{CwT}): Refers to including the  temperature into the centralized type.
\end{itemize}
The improvement in the measurements among the 4 estimation types will be compared to select the best way to improve the F/T sensor performance. For comparison, results using the Workbench matrix are included among the estimation types tables.
\section{EXPERIMENTS} 
\label{sec:experiments}
\subsection{Experimental Platform}
\begin{figure}
 \begin{minipage}[c]{0.16\textwidth}
 
    \includegraphics[width=\textwidth]{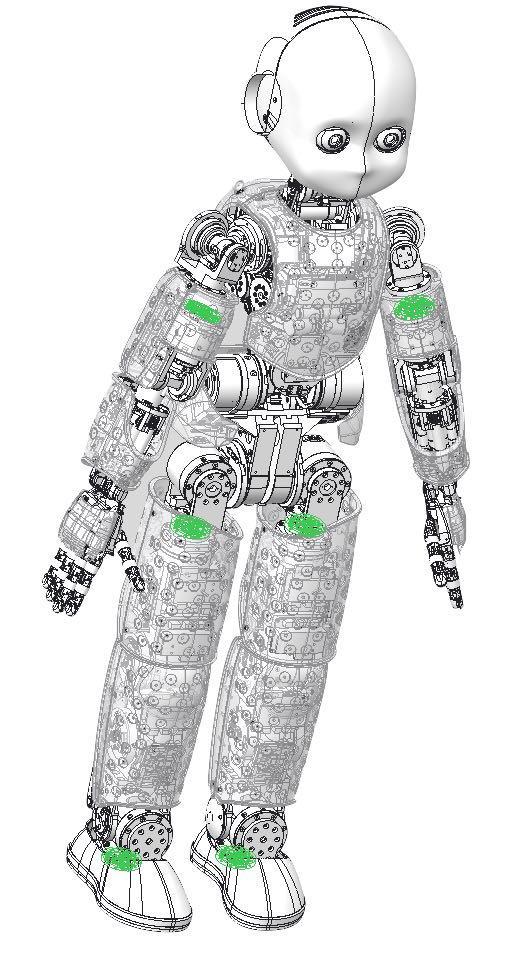}
        \caption{ The six axis F/T sensors location on the iCub.}
  \label{fig:sensors}
  \end{minipage}\hfill
\begin{minipage}[c]{0.33\textwidth}
\centering
\captionsetup{type=table}
    \rowcolors{1}{lightgray}{white}
\begin{tabular}{l|l|l|}
\cline{2-3}
 & \multicolumn{2}{l|}{Temperature Celsius} \\ \hline
\multicolumn{1}{|l|}{Data set type} & Start & End \\ \hline
\multicolumn{1}{|l|}{Grid} & 32$^o$ & 41.2$^o$ \\ \hline
\multicolumn{1}{|l|}{Balancing right} & 38.1$^o$ & 41.6$^o$ \\ \hline
\multicolumn{1}{|l|}{Validation set} & 39$^o$ & 40.5$^o$ \\ \hline
\end{tabular}\caption{Temperature values of the data sets used.}\label{tab:tempValues}
  \end{minipage}
   \end{figure}
Experiments have been performed on the 53 DOF robot iCub. It has 6 custom-made six axes F/T sensors~\cite{IITsensors} (one per ankle, leg and arm) placed as shown in Fig.~\ref{fig:sensors}. The F/T sensors mounted on the iCub use silicon strain gauge technology. In the new FTsense, the location of the temperature sensor is as close as possible to the strain gauges, making it a reliable source of temperature information. The range of temperatures observed due to normal use of the robot go from 28$^o$ Celsius to 50$^o$ Celsius. Looking at data collected during the use of the robot, it was observed that the sensors at the ankle and the arms suffer from less variation of temperature compared to the ones at the hip. Unless an specific source of heat is used to change the temperature of the robot, as in Fig. \ref{fig:temperatureEffect}, the main source of heat seems to come from the motors. For this reasons, during the experiments presented, we focus on the calibration of the F/T sensors located on the hip of the robot.
\subsection{ Types of data sets}
There are mainly 3 types of data sets used for either experiment, validation or both:
 \begin{itemize}
      \item \textbf{Grid}: moving the legs in a grid pattern on a fixed pole. The contact is on the waist of the robot. The leg is never bent so the center of mass of the leg during the experiment does not change.
      \item \textbf{Balancing}: doing an extended one foot balancing demo with widespread leg movements. The contact is on the support leg foot. Either left or right depending on the support leg.
      \item \textbf{Random}: doing random leg movements while the robot is on a fixed pole.
  \end{itemize}
The balancing movements of the Balancing data sets are the ones seen in \cite{dynVideo}.
In the data sets used for estimating the calibration matrix, it is assumed that there is only one contact point and that it is known. Other assumptions are: the drift is mainly caused by temperature, temperature drift has linear behavior and the offset is constant.
\begin{figure}[hbt!]
\centering
      \includegraphics[width=\linewidth]{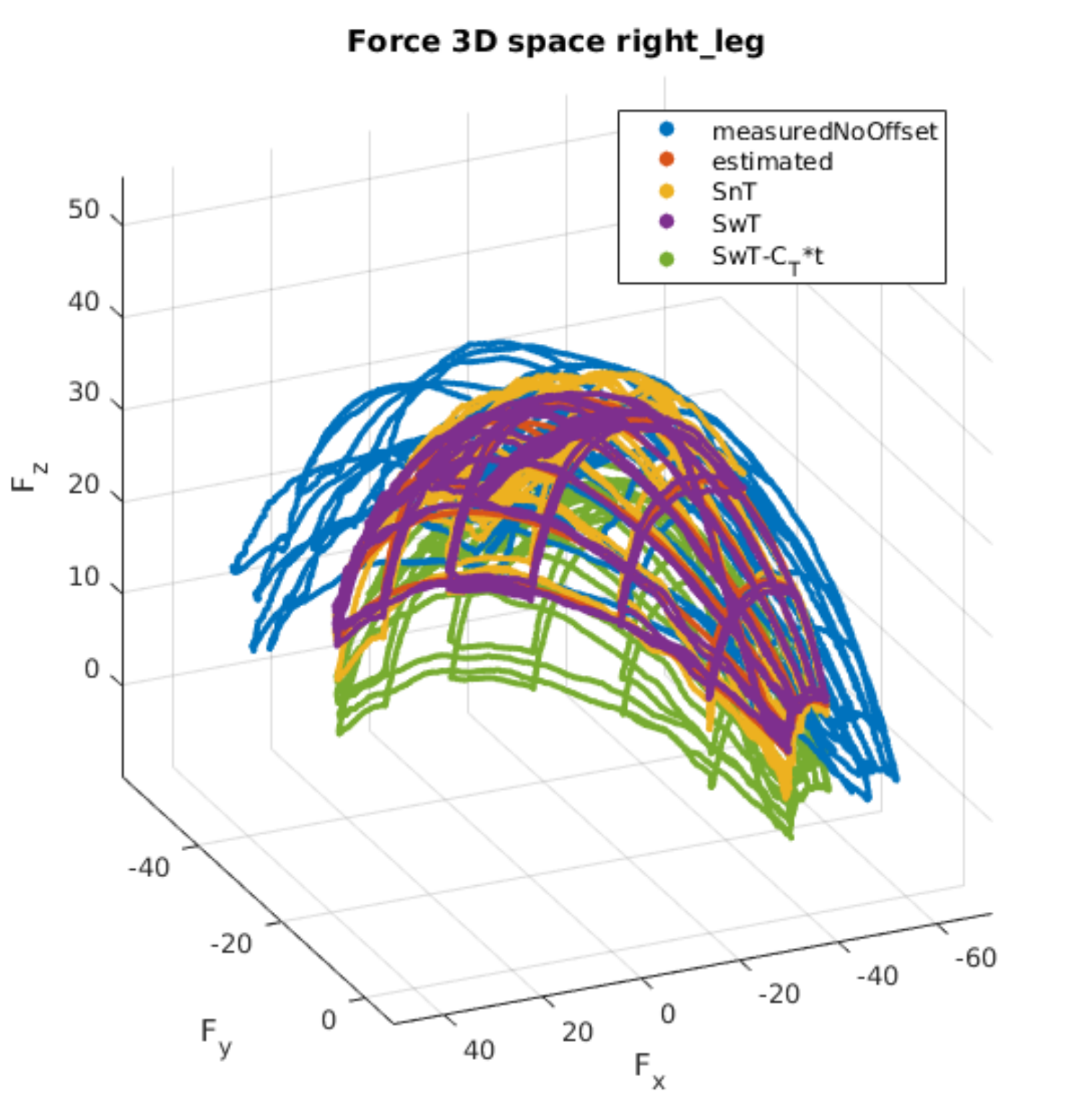}
        \caption{Re-calibrated measurements using the new estimated calibration matrices.}
         \label{fig:grid_sphere}
\end{figure}
\subsection{Experiment Description} 
All data sets were collected the same day. 3 sets of each type were collected at different temperatures. The validation set is composed of the second of each type of data set. They were collected one after the other to see the temperature change during a "normal short" session of robot use. The others are used for calibration. It is important to mention that during the use of the robot the temperature rises from 28$^o$ Celsius to 36$^o$ Celsius in a couple of minutes, but afterwards the rate of change in the temperature slows down. It takes around 2 hours of constant use to go up 50$^o$ Celsius. The temperature of the data sets used are shown in table \ref{tab:tempValues}. From previous experiments \cite{insituFTcalibration}, it was seen that the grid type of data set with sphere type of estimation gave the best results. Because of this, the data sets used to calibrate the sensor were initially formed by 2 data sets of the same type with different temperature, thus becoming a calibration data set. But, given that the temperature is a slow phenomenon, a new calibration data set was created from the combination of the grid and balancing on the right data sets. This allows to include a wider range of temperature and forces into the estimation. In the sphere type of estimation, the offset is estimated only in the first grid data set.
\subsection{ Validation procedures}
There are two main validations procedures that were used. The first is to evaluate the impact of including temperature using the method described in subsection \ref{sec:addLinearVariable}. For this, $\lambda$ is set to 0 and the Mean Square Error (MSE) of each axis was taken as performance index. The lower the value the better. This allows us to have a first insight in the improvement of using temperature or not. To make the improvement more clear we can consider the $\%$ of error reduction calculated as:
\begin{equation}
    MSE_{\%}=\left(\frac{MSE_{noT}-MSE_t}{MSE_{noT}}\right)*100,
\end{equation}
where $MSE_{noT}$ is the MSE of a estimation type without temperature, $MSE_t$ is the MSE of a estimation type with temperature and $MSE_{\%}$ is the percentage of error decrease comparing $MSE_t$ with $MSE_{noT}$ for a given calibration data set. In this case, higher error reduction equals better performance.
The second validation procedure consists on estimating the external forces in the section between the ft sensor at the hip and the one at the ankle. Since no force is exerted on the robot, the value should be 0. The 4 estimation types and 13 values of $\lambda$ [ 0, 1, 5, 10, 50, 100, 1000, 5000, 10000, 50000, 100000, 5e+05, 1e+06] were considered, which adds up to 52 calibration matrices per calibration data set. This evaluation can be performed on the magnitude of the force and in each axis. The algorithm used to estimate the external forces can be found in \cite{silvioThesis}. The external 6D force value is estimated at a given contact point. Then is brought back to the sensor frame so that a external force value in an axis matches the axis in the sensor.
\section{Results}
\label{sec:results}
\subsection{Temperature vs No Temperature} \label{sec:tempvsNotemp}
The improvement obtained by including the temperature can be seen in Fig. \ref{fig:grid_sphere}. It shows the fitting of the sphere estimation types in the grid calibration data set. The contribution of the temperature in SwT can be appreciated by looking at the distance between the purple and the green graphs.
\rowcolors{1}{lightgray}{white}
 \begin{table}[hbt!]
 \centering
\begin{tabular}{|l|l|l|l|l|}
\hline
\multicolumn{2}{|l|}{Estimation types in a data set} & \multicolumn{3}{c|}{Mean square Error (MSE)} \\ \hline
estimation type & data set & $F_{x}$ & $F_{y}$ & $F_{z}$ \\ \hline
SnT & Grid & 0.5848 & 0.3554 & 6.1136 \\ \hline
SwT & Grid & 0.5677 & 0.3553 & 1.7123 \\ \hline
CnT & Grid & 0.5803 & 0.3554 & 3.1415 \\ \hline
CwT & Grid & 0.5802 & 0.3554 & 3.1444 \\ \hline
Workbench & Grid & 37.0273 & 45.2390 & 11.8247 \\ \hline
SnT & Combined & 3.7293 & 1.7468 & 21.5244 \\ \hline
SwT & Combined & 2.4559 & 1.3176 & 6.6869 \\ \hline
CnT & Combined & 2.7165 & 1.3355 & 10.0594 \\ \hline
CwT & Combined & 2.6852 & 1.3371 & 10.3937 \\ \hline
Workbench & Combined & 47.1391 & 47.9539 & 26.2718 \\ \hline
\end{tabular}
\caption{MSE of the forces.}\label{tab:tempVsNoTempForce}
\end{table}
The values of the MSE of each estimation type in each calibration data set are shown in table \ref{tab:tempVsNoTempForce}.  The improvement of including temperature is more evident in the combined data set results than in the grid data set. In the grid data set, the highest improvement is on the z axis with a $MSE_{\%} = 71\%$, but the $MSE_{\%}$ is not as big in the other axis. Instead in the combined data set all force axis have a $MSE_{\%}$ of at least 24.5$\%$. The highest $MSE_{\%}$ is again on the z axis with 68.9$\%$. Is worth to notice that the axis related to the forces, benefit more from including temperature than the torques as shown in table \ref{tab:tempVsNoTempTorque}. There is a clear difference in performance between estimation types. The sphere types benefit more from the inclusion of temperature in the estimation than the centralized types. CnT and CwT types have a very similar performance. Note that the lowest values of MSE in the forces are achieved by the SwT estimation type regardless of the calibration data set. 
\rowcolors{1}{lightgray}{white}
 \begin{table}[hbt!]
 \centering
\begin{tabular}{|l|l|l|l|l|}
\hline
\multicolumn{2}{|l|}{Estimation types in a data set} & \multicolumn{3}{l|}{Mean square Error (MSE)} \\ \hline
estimation type & data set & $\tau_{x}$ & $\tau_{y}$ & $\tau_{z}$ \\ \hline
SnT & Grid & 17.31e-03 & 27.91e-03 & 19.30e-06 \\ \hline
SwT & Grid & 17.33e-03 & 27.61e-03 & 19.19e-06 \\ \hline
CnT & Grid & 17.32e-03 & 27.89e-03 & 18.85e-06 \\ \hline
CwT & Grid & 17.32e-03 & 27.88e-03 & 18.87e-06 \\ \hline
Workbench & Grid & 31.84e-03 & 262.31e-03 & 1.88e-03 \\ \hline
SnT & Combined & 138.09e-03 & 77.78e-03 & 14.06e-03 \\ \hline
SwT & Combined & 130.76e-03 & 63.18e-03 & 14.12e-03 \\ \hline
CnT & Combined & 131.98e-03 & 58.37e-03 & 14.13e03 \\ \hline
CwT & Combined & 131.93e-03 &58.17e-03 & 14.19e-03 \\ \hline
Workbench & Combined & 311.11e-03 &493.53e-03 & 17.99e-03 \\ \hline
\end{tabular}
\caption{MSE of the torques.}\label{tab:tempVsNoTempTorque}
\end{table}
\begin{table*}[!bhtp]
    \centering
 \begin{tabular}{|l|l|l|l|l|l|l|l|l|l|l|l|l|l|}
\hline
 & \multicolumn{13}{c|}{$\lambda$ values used} \\ \hhline{|~|-|-|-|-|-|-|-|-|-|-|-|-|-|}
\multirow{-2}{*}{Data + Estimation Type} & 0 & 1 & 5 & 10 & 50 & 100 & 1000 & 5000 & 10000 & 50000 & 1e+05 & 5e+05 & 1e+06 \\ \hline
grid SnT & 5.75 & 5.73 & 5.73 & 5.80 & 6.75 & 8.13 & 15.07 & 13.74 & 11.62 & 7.71 & 7.10 & 6.79 & 6.80 \\ \hline
grid SwT & 6.21 & 6.18 & 6.14 & 6.17 & 6.77 & 7.82 & 13.40 & 12.19 & 10.86 & 8.92 & 8.72 & 8.87 & 9.19 \\ \hline
grid CnT & 5.64 & 5.62 & 5.61 & 5.67 & 6.57 & 7.91 & 14.52 & 12.81 & 10.69 & 7.05 & 6.54 & 6.63 & 7.03 \\ \hline
grid CwT & 5.64 & 5.62 & 5.61 & 5.67 & 6.56 & 7.89 & 14.44 & 12.64 & 10.45 & 6.79 & 6.28 & 6.35 & 6.74 \\ \hline
combined SnT & 5.24 & 5.24 & 4.27 & 4.27 & 4.25 & 4.22 & \textcolor{OliveGreen}{4.13} & 4.56 & 4.86 & 6.07 & 6.46 & 6.65 & 6.76 \\ \hline
combined SwT & 7.52 & 7.51 & 4.39 & 4.38 & 4.31 & 4.24 & \textbf{\textcolor{ForestGreen}{3.99}} & 4.53 & 4.91 & 6.19 & 6.61 & 6.89 & 7.08 \\ \hline
combined CnT & 5.89 & 5.89 & 4.56 & 4.55 & 4.50 & 4.45 & \textcolor{OliveGreen}{4.21} & 4.56 & 4.83 & 5.87 & 6.18 & 6.39 & 6.74 \\ \hline
combined CwT & 5.92 & 5.91 & 4.56 & 4.55 & 4.50 & 4.45 &  \textcolor{OliveGreen}{4.21} & 4.56 & 4.82 & 5.82 & 6.10 & 6.21 & 6.52 \\ \hline
Workbench & {\textcolor{BrickRed}{10.46}} & 10.46 & 10.46 & 10.46 & 10.46 & 10.46 & 10.46 & 10.46 & 10.46 & 10.46 & 10.46 & 10.46 & 10.46 \\ \hline
\end{tabular}
    \caption{The value of external forces depending on data set, estimation type and $\lambda$ value}
    \label{tab:extForces}
\end{table*}

\subsection{External force estimation results} \label{sec:externalForce}
\begin{figure}[hbt!]
\centering
      \includegraphics[width=\linewidth]{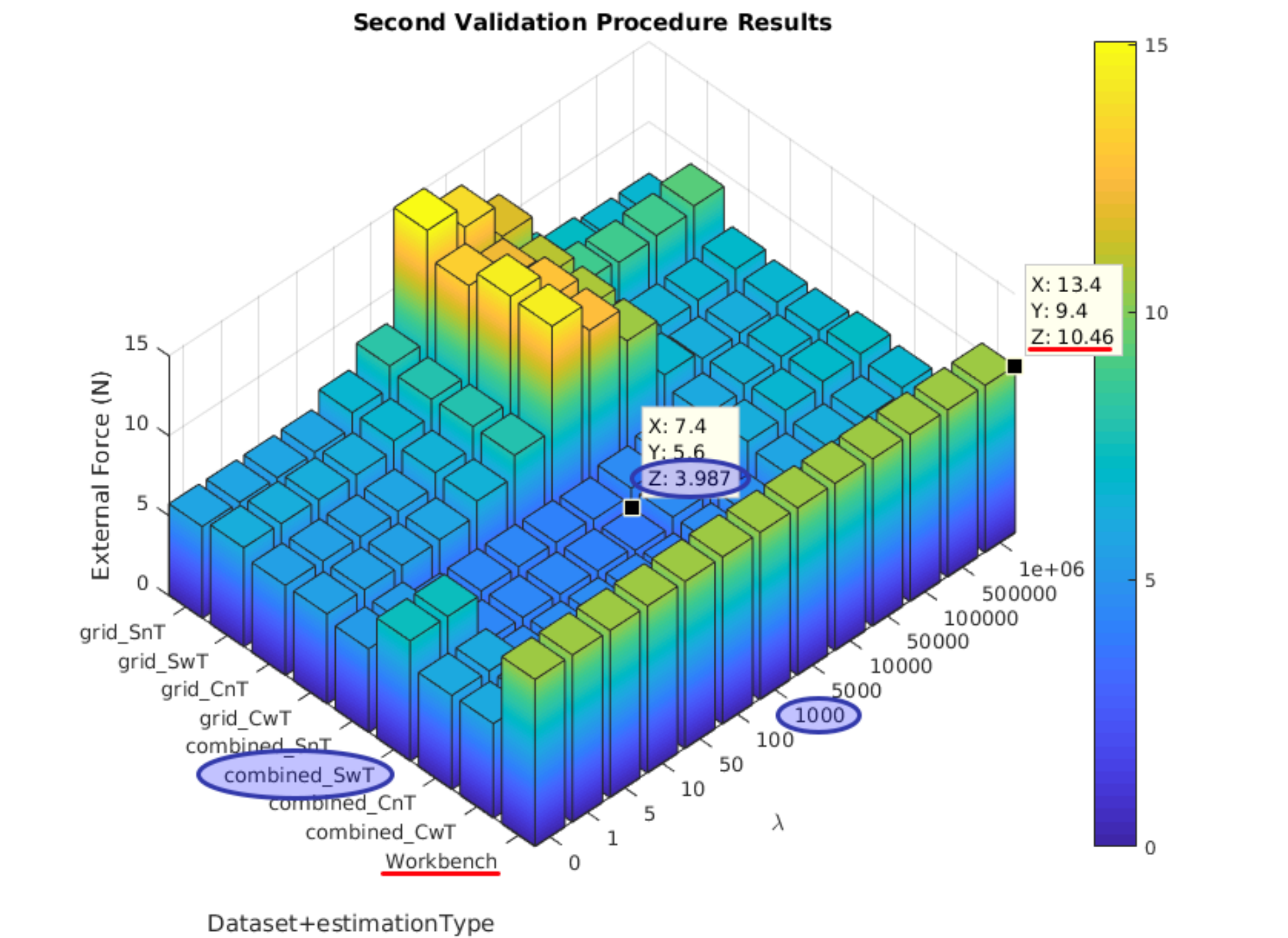}
        \caption{External Force Estimation from the proposed calibration matrices.}
         \label{fig:externalForces}
\end{figure}

 \begin{table}[bht!]
     \centering
     \begin{tabular}{c|cc|c|}
         Axis & Best $C$ & value & workbench \\
         $f_x$ & combined $\lambda$100 SwT & 2.30850 N & 5.7027 N \\
         $f_y$ & grid $\lambda$5 SwT & 1.75068 N  & 7.8396 N\\
         $f_z$ & combined $\lambda$5 SwT & 2.07207 N & 3.8282 N\\
         $\tau_x$ & combined SwT & 0.40032 Nm & 0.4406 Nm\\
         $\tau_y$ & grid $\lambda$10000 CnT & 0.41389 Nm & 0.6565 Nm \\
         $\tau_z$ & combined $\lambda$50000 CwT & 0.05992 Nm & 0.1465 Nm\\
     \end{tabular}
     \caption{Best $C$ by axis.}
     \label{tab:bestByaxis}
 \end{table}
The results of the second validation procedure are shown in Fig. \ref{fig:externalForces} and table \ref{tab:extForces}. The best result, taking all axis from the same calibration matrix, was obtained by a SwT type on the combined data set using a $\lambda$ value of 1000. The average magnitude of the external force was 3.987 N which makes it over 2 times better than the Workbench results of 10.46 N. The error decreased by 62$\%$.
It can be observed that the general performance using the combined data set is superior to using only grid data sets. The effect of the regularization is also highlighted in the combined data set. Results improve with higher penalization up to a point where being too similar to the Workbench matrix no longer gives benefits. It is curious how the grid data set seems to have an opposite behavior having better performance either with low penalization or high. In table \ref{tab:extForces}, it can be seeing that there is little advantage in doing temperature compensation using the centralized estimation type. Regardless of data set the results between using and not using temperature are very similar. Is important to notice that it still benefits from the different $\lambda$ values .
Taking a look at the results by axis in table \ref{tab:bestByaxis}, is possible to see that there is no specific $\lambda$ that gives the best result on all 6 axis. Considering that higher values of $\lambda$ means a higher similarity with the workbench, it can also give a hint on which axis are affected the most by the mounting process. From the results by axis, 5 of them improved more with the temperature compensation, 4 of them got better results with the combined data set and 4 have relatively low $\lambda$ values. The axis with the major improvement was the $f_y$ reducing the error by 77.67$\%$. The one with the least improvements was the $\tau_x$ with 9.14$\%$ error reduction.
\section{CONCLUSIONS}
\label{sec:conclusions}
It can be seen that the effect of the temperature has a relevant impact in the measurements of the F/T sensor. This is more crucial for F/T sensors expected to keep a good performance over a couple of hours of use. Because of this, robots, specially floating base robots, can benefit from temperature compensation. 
From the results in subsection \ref{sec:tempvsNotemp} it can be seen that including the temperature as a linear variable shows considerable improvement. But this improvement is also linked to the estimation type used. From the data is evident that the forces are much more affected by the temperature than the torques.
Looking at the results in subsection \ref{sec:externalForce} the following conclusions can be drawn:
\begin{itemize}
    \item Using the combined data set is better. This is probably because the sensors are exposed to a higher range of excitation values.
    \item The most successful combination of data set type and estimation type are the combined data set using SwT estimation.
    \item Including information on the previous calibration matrix can further improve the results, even if no temperature information at the moment of calibration is provided.
    \item Looking at the results by axis can help identify which axis is more affected by the mounting procedure. This could lead to a guided search in the mounting procedure to diminish the impact in the F/T sensor.
    \item Even if the centralized estimation provides less improvement than the sphere estimation, it still improves considerably with respect to the Workbench matrix.
    \item The results of the centralized estimation seem less affected by the inclusion of temperature. This makes it promising for situations where temperature information is not available and its hard to full fill the assumptions required by the sphere type of estimation.
    \item The fact that the centralized estimation is less affected by temperature, might indicate that the offset estimated through this method is not the true offset, but one mixed with the average effect of temperature. Reason for which is able to display similar results with and without temperature information. 
\end{itemize}
An advantage of the method proposed is that it can generalize to any number of linear variables, as long as enough raw data is provided. It is also able to integrate previous information from the linear variables. This allows to include other common sources of drift, such as vibration, if reliable measures of the phenomenon near the sensor are obtained.
As future work, we would like to find a performance index to see the improvement in performance of F/T controllers \cite{nori2015icub} derived from the improvement in the F/T sensor measurements. 
It is evident that putting a grid type of data set and a yoga type of data set together improves the estimation. Due to this, part of the future research will focus on finding a set of data set types that fully excite the sensor, while performing the least amount of movements. 
Considering that the temperature might be somehow included in the offset from the CnT estimation type, it will be worth performing a second \textit{in situ} estimation considering the temperature on the error between estimated data and the calibrated data without removing the offset. This might allow to decouple the true offset from the temperature effect, thus further improving the results of this estimation method.


\bibliographystyle{IEEEtran}
\bibliography{bib}

\addtolength{\textheight}{-12cm}

\end{document}